# iHHO-SMOTe: A Cleansed Approach for Handling Outliers and Reducing Noise to Improve Imbalanced Data Classification

Khaled SH. Raslan
Department of Mathematics, Faculty of Science, Al-Azhar University, Cairo, Egypt

Almohammady S. Alsharkawy
Department of Mathematics, Faculty of Science, Al-Azhar University, Cairo, Egypt

K.R. Raslan
Department of Mathematics, Faculty of Science, Al-Azhar University, Cairo, Egypt

## ABSTRACT
Classifying imbalanced datasets remains a significant challenge in machine learning, particularly with big data where instances are unevenly distributed among classes, leading to class imbalance issues that impact classifier performance. While Synthetic Minority Over-sampling Technique (SMOTE) addresses this challenge by generating new instances for the under-represented minority class, it faces obstacles in the form of noise and outliers during the creation of new samples. In this paper, a proposed approach, iHHO-SMOTe, which addresses the limitations of SMOTE by first cleansing the data from noise points. This process involves employing feature selection using a random forest to identify the most valuable features, followed by applying the Density-Based Spatial Clustering of Applications with Noise (DBSCAN) algorithm to detect outliers based on the selected features. The identified outliers from the minority classes are then removed, creating a refined dataset for subsequent oversampling using the hybrid approach called iHHO-SMOTe. The comprehensive experiments across diverse datasets demonstrate the exceptional performance of the proposed model, with an AUC score exceeding 0.99, a high G-means score of 0.99 highlighting its robustness, and an outstanding F1-score consistently exceeding 0.967. These findings collectively establish Cleansed iHHO-SMOTe as a formidable contender in addressing imbalanced datasets, focusing on noise reduction and outlier handling for improved classification models.

## Keywords
Noised Data, Data Cleansing, Imbalanced Datasets, HHO, SMOTE, DBSCAN, Random Forest.

## 1 INTRODUCTION
The pervasiveness of machine learning (ML) across diverse domains has intensified the challenge of accurately classifying imbalanced datasets. This intricate issue is further amplified by the deluge of data in the big data era [1]. In this context, the Synthetic Minority Over-sampling Technique (SMOTE) has emerged as a cornerstone technique, aiming to redress class imbalance by generating synthetic samples for the underrepresented (minority) class [2]. However, recent research has highlighted inherent limitations associated with SMOTE, particularly its susceptibility to noise and outliers during the synthetic sample creation process [3] and [4].

In modern applications, tackling the complexities of imbalanced data with noise and borderline data has become a prominent concern. This challenge is especially apparent across diverse domains, including but not limited to fraud detection in telecom, text classification, and biomedical analysis [2] and [5-7]. Delved into the theoretical distribution of SMOTE-generated data, uncovering that noise could be readily replicated during the oversampling process [3]. SMOTE sometimes focuses too much on data points near the decision boundary between classes. This can shrink the boundary and make the model too sensitive to small changes in the data, causing it to fit too closely to the training data and perform poorly on new data [8].

These limitations of SMOTE highlight the ongoing quest for more robust techniques in imbalanced learning. This paper presents an innovative approach to handle the noise data over the different data set before handling the imbalanced minority and majority classed for different datasets in different real-life application like cancer diagnosis, fraud detection, and anomaly analysis. This methodology proposed in this paper is unfolds in a comprehensive two-step procedure, beginning with thorough data cleansing to eliminate noise, followed by the application of hybrid oversampling using the HHO-SMOTe technique. By adopting this novel strategy to aim a significantly enhance the reliability and accuracy of binary classification models, especially in scenarios involving imbalanced datasets.

The DBSCAN (Density-Based Spatial Clustering of Applications with Noise) is used for eliminating the noise between the cluster as DBSCAN stands out in the data mining and machine learning world for its ability to automatically determine the number of clusters based on data density, unlike traditional methods that require pre-defined parameters (Ester et al., 1996) [9]. This eliminates user bias and adapts to the data's inherent structure. Furthermore, DBSCAN excels at identifying clusters of irregular shapes and sizes, making it robust against limitations of spherical cluster assumptions in K-Means (Ankerst et al., 1999) [10]. Its density-based approach also grants superior resilience to noise and outliers compared to other algorithms (Kriegel et al., 2015) [11]. Finally, DBSCAN operates efficiently with minimal resources, making it suitable for large datasets (Tanuja Das et al., 2024) [12]. These combined strengths make DBSCAN a versatile and powerful tool for various applications like pattern recognition, anomaly detection, and spatial data analysis.

The following sections of this paper are organized as follows: Section 2 reviews current methods for dealing with noise in imbalanced datasets. Section 3 describes in detail the algorithms used in the new method: Random Forest, DBSCAN, HHO, and SMOTE. In Section 4, explains the new iHHO-SMOTe algorithm, covering its design and how it works. Section 5 provides a detailed analysis of the results, including different datasets and algorithms. Finally, Section 6 concludes the paper.

## 2 RELATED WORK

Lot number of research work on the expatiation of the imbalanced datasets [12-16], some of them was inserted and worked on the noised and boarder lines in the imbalanced data set [17-20]. These studies not only examined current literature, but also provided various strategies to successfully address these issues in imbalanced data sets. Their aim is to identify the most optimal approach that demonstrates superior performance in addressing this issue. Xuanyu Yi [17] Introduced a noise reeducation approach for handling imbalanced datasets. They outlined the methodology and implementation steps is to train a classifier to identify noise that remains invariant across class and context distributional changes then transform hard noises into easy ones. The Approach proposed by Jinpeng Li [18] was a Multi-Environment Risk Minimization (MER) approach captures causal features between data and labels to identify noise, while the Rescaling Class-aware Gaussian Mixture Modeling (RCGM) develops detection mappings that are invariant to class variations for noise elimination. Through extensive experiments on two imbalanced noisy clinical datasets.

Huafeng Liu [19] had present Confidence-based Sample Augmentation (CSA) to bolster the reliability of selected clean samples during training, utilize prediction history to refine labels of chosen noisy samples; Additionally, introduce the Average Confidence Margin (ACM) metric, leveraging the model's evolving dynamics, to gauge the quality of corrected labels. The authors in [20] propose noise-robust designs comprising algorithm for eradicating noise within data sample clusters, adaptive embedding for secure sample generation, and a secure boundary for expanding class boundaries. the algorithm, empowered by a heterogeneous distance metric and adapted decomposition strategy, is suited for handling mixed-type and multi-class imbalanced data over benchmark datasets. The study introduced by Asniar and Nur Ulfa [21] was to use original SMOTE with enhancing its ability to detect noise within synthetic minority data generated for imbalanced data handling. By incorporating the Local Outlier Factor (LOF), SMOTE-LOF aims to improve upon the shortcomings of traditional SMOTE. The experiment involved assessing SMOTE-LOF's performance using imbalanced datasets, with comparisons drawn against SMOTE. The findings indicate that SMOTE-LOF exhibits superior accuracy and f-measure compared to SMOTE, this pre-processing step contributes significantly to enhancing the distribution of the samples.

The study introduces by KYOUNGOK [22] showing hybrid sampling\/ensemble techniques, NASBoost and NASBagging, derived from an adaptation of SMOTE. These methods mitigate the selection of noise samples from the minority class while ensuring diversity across training sets. By incorporating new criteria, the sampling approach identifies and excludes potential sources of noisy synthetic samples during SMOTE processing, which enhances classification accuracy by averting noise generation while maintaining equitable selection of minority class samples. The study suggested by M. Revathi and D. Ramyachitra [23] is to merge Noise reduction and oversampling techniques, focusing on fortifying the minority class at the borderlines. The method was on datasets with varying degrees of imbalance, also Different classifiers including Decision Tree, Gaussian Naive Bayes, Random Forest, SVM, etc were employed for validation through experiments. Which offer more accurate results in handling imbalanced datasets.

In this paper, A new method based on SMOTE was proposed, supported with machine learning algorithms, this method is designed to better tackle the challenge of eliminating noise from imbalanced datasets during synthesis. By incorporating a Random Forest Classifier for feature selection and DBSCAN for clustering data to identify outliers and noisy data, the method aims to prevent the generation of noisy data in minority/majority instances, thereby enhancing the efficiency of the classification process.

## 3 PRELIMINARIES

This section delves into the utilization of four algorithms aimed at processing imbalanced datasets; Random Forest classification, DBSCAN clustering, SMOTE oversampling, and the Harris Hawk Optimization (HHO) algorithm, encompassing with all phases. Showing comprehensive insights into the functionalities of each algorithm are provided along with their distinctive procedures. Subsequently, the integration of these algorithms into an innovative framework tailored for noise management within imbalanced datasets is showcased later on. This unified approach harnesses the unique capabilities of each algorithm to flexibly adjust to diverse datasets, consistently delivering superior results with remarkable efficiency.

### 3.1 Random Forest

Random Forest (RF) has emerged as a prominent ensemble learning technique for classification and regression tasks due to its high predictive performance and robustness against overfitting. An integral aspect of RF's efficacy lies in its ability to perform feature selection during the model training process. This section delves into the inner workings of RF, elucidating how it selects features and estimates their importance in the context of building predictive models. It works in 2 important stage which are Random Forest Feature Selection, Estimation of Feature Importance that will be explain in the below sections.

### 3.1.1 Random Forest Feature Selection

In RF, feature selection is implicitly performed during the construction of individual decision trees, which collectively constitute the ensemble model. At each node split of a decision tree, a random subset of features is considered for determining the best split. This randomness injects diversity into the trees, thereby reducing the correlation among them and mitigating overfitting [24].

Let $X$ denote the input feature matrix with $n$ samples and $p$ features, and $Y$ denote the corresponding response variable. The RF algorithm entails the following steps:

1. Randomly select a subset of features of size $m$ ($m \ll p$) at each node split.
2. For classification tasks, the Gini impurity or information gain criterion is used to find the optimal split among the selected features. For regression tasks, the mean squared error reduction criterion is employed.
3. Grow multiple decision trees using bootstrapped samples from the training data.
4. Aggregate the predictions of individual trees to obtain the final prediction through averaging (for regression) or voting (for classification).

### 3.1.2 Estimation of Feature Importance

Random Forest determines feature importance by evaluating the decrease in impurity or information gain brought about by each feature during the tree-building process. For classification tasks, the Gini impurity index is commonly utilized, while for regression tasks, the mean decrease in impurity or variance is employed. The importance of a feature $X_i$ can be calculated as:

$$Importance(X_i) = \sum_{t=1}^{T} \frac{n_t}{n} \left( impurity_{parent} - impurity_{child} \right) \quad (1)$$

Where $T$ denotes the number of trees in the forest, $n_t$ represents the number of samples reaching node $t$, $n$ signifies the total number of samples, and $impurity_{parent}$ and $impurity_{child}$ denote the impurity measures of the parent and child nodes, respectively.

Random Forest's feature selection mechanism was used because it has several advantages; Firstly, it can handle a large number of features without overfitting, thereby accommodating high-dimensional datasets. Secondly, it provides robustness against noise and multicollinearity, as the ensemble nature mitigates the impact of individual trees' biases. Moreover, RF inherently captures interactions among features, enhancing the interpretability of selected variables [24] and [25].

It's crucial to note that while Random Forest is effective for feature selection, its performance might vary depending on the dataset characteristics, such as class imbalance and feature distributions. Additionally, the choice of hyperparameters, such as the number of trees and the splitting criterion, can influence feature importance scores. Hence, careful parameter tuning, and cross-validation are recommended to ensure optimal feature selection performance [26].

## 3.2 DBSCAN Clustering and Outlier Detection

Density-Based Spatial Clustering of Applications with Noise (DBSCAN) stands as a robust algorithm renowned for its ability to cluster spatial datasets based on the density of data points. Unlike conventional clustering techniques, such as k-means, DBSCAN operates without requiring the a priori specification of the number of clusters, rendering it particularly useful for datasets with irregular shapes and varying densities [27].

DBSCAN's fundamental concept revolves around the notion of data point density. Introduced by Ester et al. in 1996, the algorithm identifies core points, border points, and noise points (outliers) based on two primary parameters: ε (epsilon): A radius within which to search for neighboring points and MinPts: The minimum number of points required to form a dense region [9].

Given a dataset $X = \{x_1, x_2, ..., x_n\}$, DBSCAN categorizes each point into one of three categories: (1). Core points: Points that have at least MinPts points (including itself) within distance ε. (2). Border points: Points that are within distance ε of a core point but do not have enough neighbors to be considered a core point themselves. (3). Noise points: Points that lack sufficient neighbors to be classified as either core or border points.

### 3.2.1 Overview of DBSCAN Algorithm

The DBSCAN algorithm iteratively explores the dataset, expanding clusters by incorporating reachable points. A point is considered reachable from another if there exists a path of core points connecting the two. Through this process, DBSCAN efficiently identifies clusters of varying shapes and sizes, adapting dynamically to the data distribution. The DBSCAN algorithm work with iterative manner which is consists of three main steps as below:

Core Point Identification: For each point $x_i$ in the dataset, determine if it is a core point by counting the number of points within distance ε. If $"NumPts" (x\_i) > "MinPts"$, mark $x_i$ as a core point.

Cluster Expansion: For each core point $x_i$, recursively expand a cluster by including all points that are reachable from $x_i$ within distance ε. A point $x_j$ is considered reachable from $x_i$ if there exists a path of core points $(x_1, x_2, ..., x_m)$ such that $x_1 = x_i$ and $x_m = x_j$, and each consecutive pair $x_k, x_k + 1$ in the path is within distance ε of each other. Cluster Assignment: Assign each border point to the cluster of its nearest core point. Noise points remain unassigned.

DBSCAN has a time complexity of $O(nlogn)$, making it suitable for large datasets. Furthermore, it does not require specifying the number of clusters beforehand and can handle clusters of arbitrary shapes [27] and [28].

### 3.2.2 Outlier Detection using DBSCAN

One of the distinctive features of DBSCAN is its capacity to discern outliers, or noise points, as a natural byproduct of the clustering process. Outliers, often indicative of anomalous instances or data errors, are isolated points or regions characterized by low density compared to their surroundings [28]. DBSCAN identifies such outliers by flagging points that fail to meet the criteria for core or border points, thereby providing valuable insights into data quality and anomalous behavior. By designating such points as outliers, DBSCAN assists in identifying instances that may require further investigation.

The detection of outliers using DBSCAN relies on the notion of core points and their connectivity within dense regions. Outliers are often characterized by their isolation from dense clusters, resulting in fewer neighboring points within the defined ε-radius. Consequently, noise points that fail to meet the criteria for core or border points are flagged as outliers by the algorithm. Mathematically, the identification of outliers in DBSCAN can be expressed as follows:

A point $p$ is considered a core point if:

$$|q \in D : dist(p,q) \leq \varepsilon| \geq \text{MinPts} \quad (2)$$

Where $dist(p,q)$ denotes the distance between points $p$ and $q$. A point $p$ is labeled as an outlier if it is not a core point and does not fall within the neighborhood of any core point.

DBSCAN's outlier detection mechanism has been pivotal in various domains, including fraud detection, network security, and environmental monitoring. By robustly capturing the underlying data distribution, DBSCAN facilitates the identification of subtle anomalies that may evade traditional detection methods [29]. Over the years, DBSCAN has witnessed numerous extensions and adaptations tailored to specific applications and datasets. One notable extension is the OPTICS algorithm, which extends the concepts of DBSCAN to provide a hierarchical clustering structure without requiring the predefinition of ε [10] and [30]. This hierarchical approach enables the identification of clusters at multiple scales, offering deeper insights into the data structure.

## 3.3 SMOTE

SMOTE is commonly used when dealing with imbalanced datasets, where one class (minority class) has significantly fewer examples than the other class (majority class). In such cases, machine learning models may struggle to correctly classify the minority class because they tend to be biased towards the majority class. SMOTE helps address this imbalance by generating synthetic examples of the minority class to create a more balanced dataset for training.

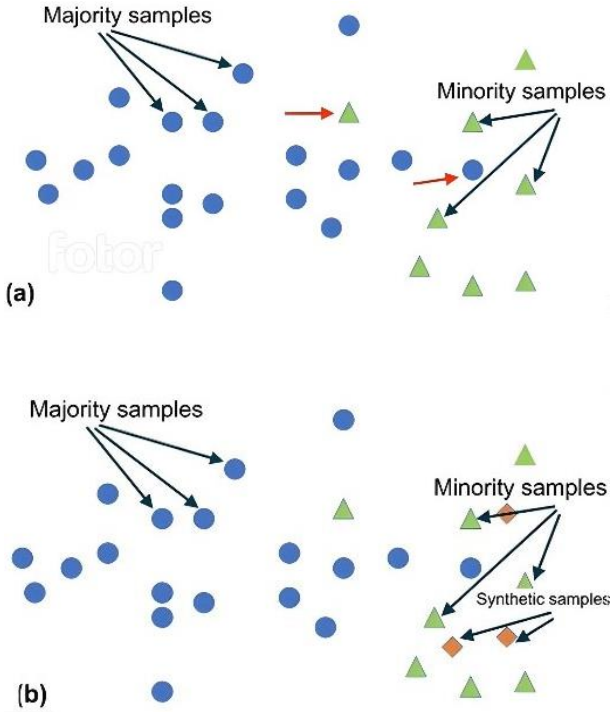

**Fig 1: The principle of the SMOTE**

An example of an imbalanced dataset can be observed in Fig.1(a) above. Here, the majority class is represented by circular shapes, which stand in for the data's predominant occurrences, while the minority class is represented by triangular shapes, signifying the smaller number of data samples. Some examples from the minority and majority classes are in areas that do not naturally align with the opposite class, most notably with the red arrow. The SMOTE algorithm initiates the process of selecting synthetic samples, a crucial step in bolstering the minority class. The sampling rate specified for each category of occurrences serves as the basis for this selection process. The synthetic samples are presented as square forms in Fig.1(b). Upon applying the SMOTE technique, the resultant effect is a reduction in the disparity between the Minority and Majority classes.

The SMOTE algorithm includes a sample rate parameter to control the extent of over-sampling. The sample rate determines how many synthetic examples are generated for each minority class instance. Here's an equation that includes the sample rate in the SMOTE algorithm:

$$C = A + s * (B - A) \quad (3)$$

Where: $C$ is the synthetic example being generated. $A$ is a randomly selected instance from the minority class. $B$ is one of the $k$ nearest neighbors of $A$ (also from the minority class) and is randomly chosen. $s$ is the sample rate parameter, which influences how many synthetic examples are generated between $A$ and $B$. The $s$ parameter is a value between 0 and 1, which allows you to control the density of synthetic examples to be generated. When $s = 0.5$, one synthetic example is generated exactly. This can be seen as an average or balanced interpolation between the two instances. If $s$ is less than 0.5, the synthetic examples will be closer to $A$ than $B$, otherwise the synthetic examples will be closer to $B$ than $A$.

## 3.4 Harris Hawks Optimizer (HHO)

The HHO has introduced by Ali Asghar Heidari in 2019, the HHO algorithm has garnered significant attention from the research community [31] and [32]. HHO draws inspiration from the hunting behavior of Harris Hawks in nature, particularly their agile surprise pounce technique. Harris Hawks, known for their remarkable intelligence, exhibit various chasing styles based on different scenarios and the behavior of their prey. HHO is widely recognized as one of the most effective optimization algorithms, and it has been successfully applied to a variety of problems across different domains encompass energy and power flow analysis, engineering, medical applications, network optimization, and image processing. The comprehensive review [33] and [34] presents a survey of the existing body of work related to HHO.

Within this section, shows the modeling of both the exploratory and exploitative phases inherent in HHO methodology. The phases are done by three steps draw inspiration from the natural behaviors of Harris hawks, including their approaches to prey exploration, surprise pouncing, and the diverse attack strategies employed. HHO represents a population-based optimization approach devoid of gradients, rendering it adaptable to a wide array of optimization challenges, provided that they are appropriately formulated. The detailed explanations provided in the subsequent subsections.

### 3.4.1 Exploration phase

Hawks perch in specific locations and constantly monitor the surrounding environment to identify prey using two strategies, which are represented in Eq.4. If $p < 0.5$, the hawks perch based on the position of the family members. If $p \geq 0.5$, the hawks perch in a random space within the population area.

$$X(t+1) = \begin{cases} X_{rand}(t) - r_1|X_{rand}(t) - 2r_2 X(t)| & q \geq 0.5 \\ (X_{rabbit}(t) - X_m(t)) - r_3(LB + r_4(UB - LB)) & q < 0.5 \end{cases} \quad (4)$$

Where $X(t+1)$ is the position vector of hawks in the next iteration $t$, $X_rabbit(t)$ is the position of rabbit, $X(t)$ is the current position vector of hawks, $r_1, r_2, r_3, r_4$, and $q$ are random numbers inside (0,1), which are updated in each iteration, LB and UB show the upper and lower bounds of variables, $X_{rand}$ is a randomly selected hawk from the current population, and $X_m$ is the average position of the current population of hawks.

The HHO utilized a simple model to generate random locations inside the group's home range (LB, UB). The first rule generates solutions based on a random location and other hawks. In second rule of Eq.4, it has the difference of the location of best so far and the average position of the group plus a randomly-scaled component based on range of variables, while $r_3$ is a scaling coefficient to further increase the random nature of rule once $r_4$ takes close values to 1 and similar distribution patterns may occur. Utilizing the simplest rule, which can mimic the behaviors of hawks. The average position of hawks is attained using Eq.5:

$$X_m(t) = \frac{1}{N} \sum_{i=1}^{N} X_i(t) \quad (5)$$

Where $X_i(t)$ indicates the location of each hawk in iteration $t$ and $N$ denotes the total number of hawks.

### 3.4.2 Transition from exploration to exploitation

The HHO can transfer from exploration to exploitation and then, change between different exploitative behaviors based on the escaping energy of the prey. The energy of a prey decreases considerably during the escaping behavior. To model this fact, the energy of a prey is modeled as:

$$E = 2E_0 \left(1 - \frac{t}{T}\right) \quad (6)$$

Where E indicates the escaping energy of the prey, $T$ is the maximum number of iterations, and $E_0$ is the initial state of its energy. In HHO, $E_0$ randomly changes inside the interval $(-1,1)$ at each iteration. When the value of $E_0$ decreases from 0 to -1, the rabbit is physically flagging, whilst when the value of $E_0$ increases from 0 to 1, it means that the rabbit is strengthening.

### 3.4.3 Exploitation phase

Which the hawks attack the targeted prey. then, however, the prey tries to escape the attack. Based on hawk attacking behavior and escaping prey behavior, four scenarios will be described as below:

**Soft besiege:**

When $r \geq 0.5$ and $|E| \geq 0.5$, the rabbit still has enough energy, and try to escape by some random misleading jumps but finally it cannot. During these attempts, the Harris' hawks encircle it softly to make the rabbit more exhausted and then perform the surprise pounce. This behavior is modeled by the following rules:

$$X(t+1) = \Delta X(t) - E|JX_rabbit(t) - X(t)| \quad (7)$$

$$\Delta X(t) = X_rabbit(t) - X(t) \quad (8)$$

Where $\Delta X(t)$ is the difference between the position vector of the rabbit and the current location in iteration $t$, $r_5$ is a random number inside $(0,1)$, and $J = 2(1 - r_5)$ represents the random jump strength of the rabbit throughout the escaping procedure. The $J$ value changes randomly in each iteration to simulate the nature of rabbit motions.

**Hard besiege:**

When $r \geq 0.5$ and $|E| < 0.5$, the prey is so exhausted, and it has a low escaping energy. In addition, the Harris' hawks hardly encircle the intended prey to finally perform the surprise pounce. In this situation, the current positions are updated using:

$$X(t+1) = X_rabbit(t) - |\Delta X(t)| \quad (9)$$

**Soft besiege with progressive rapid dives:**

When still $|E| \geq 0.5$ but $r < 0.5$, the rabbit has enough energy to successfully escape and still a soft besiege is constructed before the surprise pounce. This procedure is more intelligent than the previous case, the final strategy for updating the positions of hawks in the soft besiege phase can be performed by:

$$X(t+1) = \begin{cases} Y \, if \, F(Y) < F(X(t)) \\ Z \, if \, F(Z) < F(X(t)) \end{cases} \quad (10)$$

here $Y$ and $Z$ are obtained using Eq.11 and Eq.12. A simple illustration of this step for one hawk. $Y$ is the hawks next move based on the following rule

$$Y = X_rabbit(t) - E|JX_rabbit(t) - X(t)| \quad (11)$$

To mathematically model the escaping patterns of the prey and leapfrog movements (as called in [22]), the levy flight (LF) concept is utilized in the HHO algorithm. In HHO the hawks dive based on the LF-based patterns using the following rule

$$Z = Y + S \times LF(D) \quad (12)$$

Where $D$ is the dimension of problem and $S$ is a random vector by size 1 x D and LF is the levy flight function, which is calculated as follows.

$$LF(X) = 0.01 \times \frac{u \times \sigma}{|v|^{\frac{1}{\beta}}}, \sigma = \left(\frac{\Gamma(1+\beta) \times sin\left(\frac{\pi\beta}{2}\right)}{\Gamma\left(\frac{1+\beta}{2}\right) \times \beta \times 2^{\frac{\beta-1}{2}}}\right)^{\frac{1}{\beta}} \quad (13)$$

Where $u$, $v$ are random values inside $(0,1)$, $\beta$ is a default constant set to 1.5.

**Hard besiege with progressive rapid dives:**

When $|E| < 0.5$ and $r < 0.5$, the rabbit has not enough energy to escape and a hard besiege is constructed before the surprise pounce to catch and kill the prey. The situation of this step in the prey side is similar to that in the soft besiege, but this time, the hawks try to decrease the distance of their average location with the escaping prey. Therefore, the following rule is performed in hard besiege condition:

$$X(t+1) = \begin{cases} Y \, if \, F(Y) < F(X(t)) \\ Z \, if \, F(Z) < F(X(t)) \end{cases} \quad (14)$$

Where Y and Z are obtained using rules in Eq.15 and Eq.16.

$$Y = X_rabbit(t) - E|JX_rabbit(t) - X_m(t)| \quad (15)$$

$$Z = Y + S \times LF(D) \quad (16)$$

## 4 THE PROPOSED ALGORITHM

In this section, introducing the proposed approach for efficiently clearing datasets of noise data before creating samples. This process also involves applying the HHO algorithm to select the optimal sample rate to be used in the SMOTE. The primary goal of this hybrid algorithm is to improve the accuracy of classification for imbalanced and noisy datasets.

Firstly, employing the Random Forest algorithm for feature selection. Random Forest, an ensemble learning method, constructs multiple decision trees during training and outputs the mode of the classes for classification. This algorithm is particularly effective for feature selection because it ranks features based on their importance, which is determined by their contribution to the prediction accuracy. By selecting the most significant features, it will reduce dimensionality and focus on the most influential attributes, enhancing the subsequent clustering process.

Once the key features are identified, then apply the DBSCAN algorithm. DBSCAN is a powerful clustering method capable of identifying clusters of varying shapes and sizes, as well as detecting outliers. It operates based on two parameters: epsilon (the maximum distance between two samples for them to be considered as in the same neighborhood) and minPts (the minimum number of points required to form a dense region). DBSCAN groups together points that are closely packed and marks points that lie alone in low-density regions as outliers. This step is critical for cleansing the dataset by removing noise, which can significantly degrade the performance of machine learning models.

After the noise and outliers are identified and removed, then proceed with the HHO algorithm. HHO is a nature-inspired optimization algorithm based on the cooperative behavior and chasing strategy of Harris hawks in nature. This algorithm is used to determine the optimal sample rate for SMOTE. By simulating the chasing strategy of Harris hawks, the HHO algorithm searches for the best solution to balance the dataset. It effectively adjusts the oversampling rate to ensure that the synthetic samples generated are both meaningful and beneficial for improving the classifier's performance.

Next, integrate the optimized sample rate into the SMOTE technique. SMOTE generates synthetic samples by interpolating between existing minority class samples. The integration of HHO ensures that the sampling process is tailored to the specific needs of the dataset, avoiding overfitting and enhancing the classifier's ability to generalize from the data. The hybrid approach of combining Random Forest for feature selection, DBSCAN for clustering and outlier detection, and HHO-SMOTE for oversampling creates a robust framework for handling imbalanced datasets.

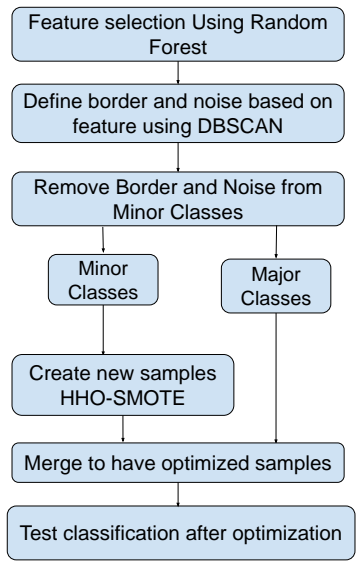

**Fig 2: Block diagram for the iHHOW-SMOTE**

The above diagram Fig.2 illustrates the integration of these algorithms. Initially, the dataset undergoes feature selection using Random Forest to identify the most relevant features. These selected features are then used in the DBSCAN algorithm to form clusters and detect outliers. The outliers are removed, resulting in a cleansed dataset. Subsequently, the HHO algorithm is applied to determine the optimal sample rate, which is then used by SMOTE to generate new synthetic samples. This entire process aims to enhance the classification results by ensuring that the new samples added to the dataset are free from noise and well-distributed across the feature space.

By adopting this comprehensive approach, aiming to address the challenges posed by imbalanced and noisy datasets. The hybrid method not only improves the balance of the dataset but also enhances the robustness and accuracy of the classification models.

**Algorithm 1 iHHO-SMOTe algorithm**

Require: $N, Fn, \varepsilon, M\ inP\ ts, T$

1: Apply random forest to extract $Fn$ from data set with size $N$
2: Cluster DBSCAN to identify noise and borderline with parameters $\varepsilon\ and\ M\ inP$ ts
3: Remove the noised and borderline from minority classes
4: **while** stopping condition is not met **do**
5: **Generate a synthetic data based on current sample rate (solution) using SMOTE alg., then calculate the fitness values of hawks using on KNN alg**.
6: **Set** Xrabbit as the location of rabbit (**highest accuracy**)
7: for each $hawk(X_i)$ do
8: Update the initial energy $E_0$ and jump strength $J \to E_0 = 2 * rand(\ ) - 1, J = 2 * (1 - rand(\ ))$
9: Update the $E$ using Eq.4
10: **if** $|E| \geq 1$ **then** (Exploration phase)
11: Update the location vector using Eq.2
12: **end if**
13: **if** $|E| < 1$ **then** (Exploitation phase)
14: **if** $r \geq 0.5$ and $|E| \geq 0.5$ **then** (Soft besiege)
15: Update the location vector using Eq.5
16: **else if** $r \geq 0.5$ and $|E| < 0.5$ **then** (Hard besiege)
17: Update the location vector using Eq.7
18: **else if** $r < 0.5$ and $|E| \geq 0.5$ **then** (Soft besiege with progressive rapid dives)
19: Update the location vector using Eq.8
20: **else if** $r < 0.5$ and $|E| < 0.5$ **then** (Hard besiege with progressive rapid dives)
21: Update the location vector using Eq.12
22: **end if**
23: **end if**
24: **end for**
25: **end while**
26: **Merge** Minority and Majority to Create **DS**
27: **Return DS**(Cleansed Dataset after apply iHHO-SMOTE)

The iHHO-SMOTe algorithm begins by taking a dataset and utilizing several key parameters such as population size, number of features, and the maximum number of iterations. Initially, it employs a random forest technique to identify and select the most significant features within the dataset, which helps in reducing dimensionality and focusing on the most relevant data points. Following this, the algorithm incorporates a clustering method known as DBSCAN (Density-Based Spatial Clustering of Applications with Noise). This method is used to detect and eliminate noise, as well as to identify and remove borderline cases within the minority class, thereby enhancing the quality and reliability of the data being analyzed.

Then the algorithm enters a phase where it iteratively processes the data. During this phase, the algorithm repeatedly runs its computations until a predefined stopping condition is satisfied. This stopping condition might be based on reaching a certain number of iterations, achieving a specific accuracy level, or other performance metrics. Throughout this iterative process, the algorithm fine-tunes its parameters and continuously seeks to improve the results.

Once the stopping condition is met, the algorithm invokes another component designed to optimize the final output as accurate and robust as possible, addressing any remaining issues and refining the model's performance. By following these steps, the iHHO-SMOTe algorithm effectively balances the dataset, improves quality of minority class data. This comprehensive approach ensures that the algorithm is capable of handling complex datasets and delivering high-quality results.

This algorithm is designed to run on a dataset DS with a population size N. The parameters involved in the algorithm are $F_n$ Number of features to be extracted using the Random Forest algorithm, $\varepsilon$ Maximum radius to search for neighboring points in the DBSCAN clustering process, $MinPts$ Minimum number of points required to form a dense region in DBSCAN, T: Maximum number of iterations for the Harris Hawks Optimization (HHO) run, the steps of IHHO-SMOTe are shown in Algorithm 1.

## 5 Performance Evaluation Measures

Performance evaluation metrics are essential for assessing how well a classification model works and for helping design better classifiers. In this analysis, a confusion matrix was used to evaluate the results of the proposed iHHO-SMOTE approach and to compare it with other SMOTE methods. The confusion matrix in Fig.3 shows how well the classification models performed. Here are some key terms: True Positive (TP) means the model predicted a positive outcome, and it was actually positive. False Positive (FP) means the model predicted a positive outcome, but it was actually negative. True Negative (TN) means the model predicted a negative outcome, and it was actually negative. False Negative (FN) means the model predicted a negative outcome, but it was actually positive. Based on the confusion matrix, several important measures can be calculated.

**Fig. 3 Confusion matrix for the two-class classification problem**

First, accuracy measures how often the model's predictions were correct. It is the proportion of true positive and true negative predictions out of all predictions made. Precision, also known as positive predictive value, is the proportion of true positive predictions out of all positive predictions made by the model. Recall, also known as sensitivity or true positive rate, is the proportion of actual positive cases that were correctly identified by the model. Specificity, or true negative rate, is the proportion of actual negative cases that were correctly identified.

Another important metric is the F1 score, which is the harmonic mean of precision and recall. This metric balances precision and recall, providing a single measure of a model's performance. The false positive rate (FPR) is the proportion of actual negative cases that were incorrectly predicted as positive. The false negative rate (FNR) is the proportion of actual positive cases that were incorrectly predicted as negative. These metrics give us a detailed understanding of a classifier's performance, highlighting different aspects of its predictive capabilities. When comparing the iHHO-SMOTE approach with other SMOTE-based approaches, these measures help determine which method strikes the best balance between true positive and false positive rates, among other factors.

**G-mean**: The G-mean, or geometric mean, is a way to measure a model's performance by looking at how well it predicts each class. It is especially useful when you want to ensure that the model performs equally well across all classes, rather than excelling in one class while failing in another [6].

For binary classification, which involves only two classes, the G-mean is calculated by taking the square root of the product of sensitivity (how well the model identifies positive cases) and specificity (how well the model identifies negative cases). This means that both sensitivity and specificity need to be high for the G-mean to be high, ensuring that the model is balanced and not biased towards one class.

When dealing with multi-class problems, where there are more than two classes, the calculation of G-mean becomes a bit more complex. Instead of the square root, you take a higher root that corresponds to the number of classes. Essentially, you multiply the sensitivity values for all the classes and then take the root based on the number of classes. This method ensures that the performance is balanced across all classes, maintaining high accuracy for each one.

$$\text{G-mean} = \sqrt{(\text{Sensitivity} * \text{Specificity})} \qquad (17)$$

**F1 score**: The F1 score, also known as the F score or F measure, is a way to measure a model's accuracy by considering two important factors: precision and sensitivity (also called recall). Precision measures how many of the positive predictions made by the model are actually correct, while sensitivity measures how well the model identifies all the actual positive cases [35].

The F1 score combines these two factors into a single number by calculating their harmonic mean. This means it gives equal importance to both precision and sensitivity, making it a balanced measure of a model's performance. If a model has high precision but low sensitivity, or vice versa, the F1 score will reflect this imbalance. By looking at the F1 score, you can get a better understanding of how well the model is performing overall, taking into account both the accuracy of its positive predictions and its ability to find all positive cases. The F1 can be calculated as:

$$F1 = \frac{2*Precision*Recall}{Precision+Recall} = \frac{2*TP}{2*TP+FP+FN} \qquad (18)$$

**AUC**: The AUC, or Area Under the Curve, is a metric used to evaluate the quality of a classification model. It is based on the ROC curve, which stands for Receiver Operating Characteristics. The ROC curve is a graph that shows the relationship between sensitivity (true positive rate) and the false positive rate for different threshold values [22].

To understand the ROC curve, imagine a graph with two axes. The X-axis represents the false positive rate, which is how often the model incorrectly predicts a positive outcome. The Y-axis represents the true positive rate, which is how often the model correctly predicts a positive outcome. By plotting these rates for various thresholds, you get the ROC curve.

The AUC is the area under this ROC curve. A larger area indicates a better performance by the model, as it shows that the model has a higher true positive rate and a lower false positive rate. Essentially, the closer the AUC is to 1, the better the model is at distinguishing between positive and negative cases. If the AUC is 0.5, it means the model is no better than random guessing. Thus, a high AUC value is desirable as it signifies a good performing model. The AUC can be calculated as:

$$AUC = (1 + TPR - FPR)/2 \qquad (19)$$

## 5.1 Experiments and Evaluation

In this section, discussion on the experiments conducted on various datasets. The upcoming parts will show the results and provide an analysis of these findings. The experiments were performed using Google Colab, with offers free Jupyter notebook environment with GPU support. This setup is ideal for running machine learning experiments [36].

In the experiment work, the new proposed method had been applied on seven datasets with different business natures and data characteristics. DS1 is the abalone9-18 dataset, which contains 8 attributes and 731 instances. DS2 is the cleveland-0 dataset, containing 177 instances. DS3 is the ecoli3 dataset, with 336 instances. DS4 is the german dataset, which has 1,000 instances. DS5 is the glass2 dataset, containing 214 instances. DS6 is the haberman dataset, with 306 instances. Finally, DS7 is the hypothyroid dataset, which contains 3,163 instances.

In Fig.4, evaluating the G-mean values for the 7 datasets using seven different SMOTE techniques. The G-mean is an important metric for assessing a model's ability to correctly identify both positive and negative classes, especially in imbalanced datasets. A higher G-mean means the model is doing a good job in this regard. Among the techniques tested, ANS-SMOTE and GASMOTE had the lowest G-mean values, indicating they were less effective. On the other hand, ADASYN, SMOTE, RANDOM-SMOTE, and Borderline-SMOTE showed similar performance, with ADASYN having a slightly lower G-mean for the "ds10" dataset.

iHHO-SMOTe stood out by consistently achieving high G-mean values across various datasets. This consistency shows that iHHO-SMOTe is very reliable for handling imbalanced classification tasks.

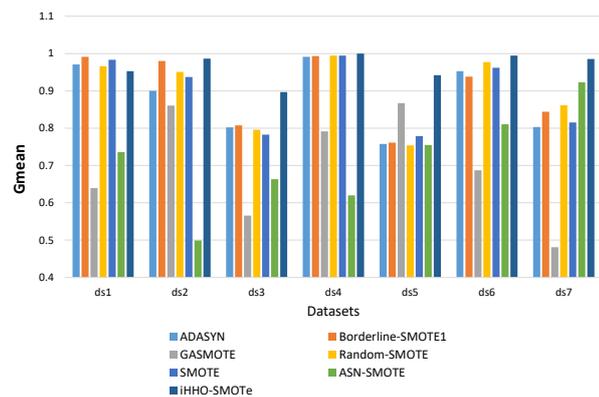

**Fig. 4 Comparison of G-mean of seven SOMTE techniques**

In Fig.5, compare the classification results of different SMOTE algorithms using the F1-score. The F1-score combines precision and recall, providing a measure of how well a model balances false positives and false negatives.

The results showed that ANS-SMOTE and GASMOTE did not perform as well as ADASYN, SMOTE, RANDOM-SMOTE, and Borderline-SMOTE. Among all the methods, iHHO-SMOTe stood out by consistently achieving high F1-scores, ranging from 0.9 to 1, across various datasets. This indicates that iHHO-SMOTe is stable and reliable for different classification tasks, effectively managing the balance between precision and recall.

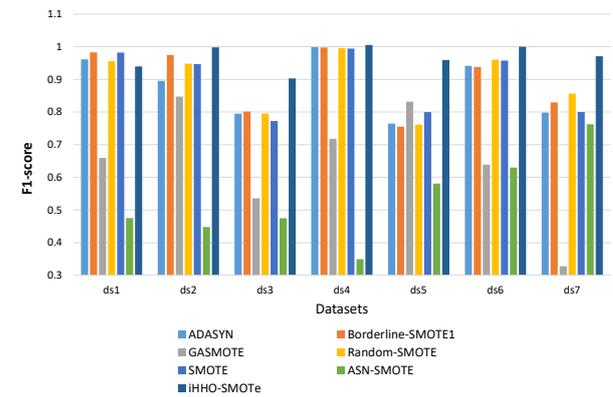

**Fig. 5 Comparison of F1-Score of seven SOMTE techniques**

In Fig.6, reevaluate the classification studies by focusing on the AUC (Area Under the Receiver Operating Characteristic Curve). AUC measures how well a binary classification model can distinguish between positive and negative cases, taking into account both true positive and false positive rates at different thresholds.

The results revealed that ANS-SMOTE and GASMOTE had lower AUC scores, indicating poorer performance. On the other hand, ADASYN, SMOTE, RANDOM-SMOTE, and Borderline-SMOTE performed better in terms of AUC. Notably, HHO-SMOTe consistently achieved high AUC values, typically ranging from 0.9 to 1. This consistency shows that HHO-SMOTe is highly adaptable across various datasets and is particularly effective in classification tasks where it is important to accurately separate classes.

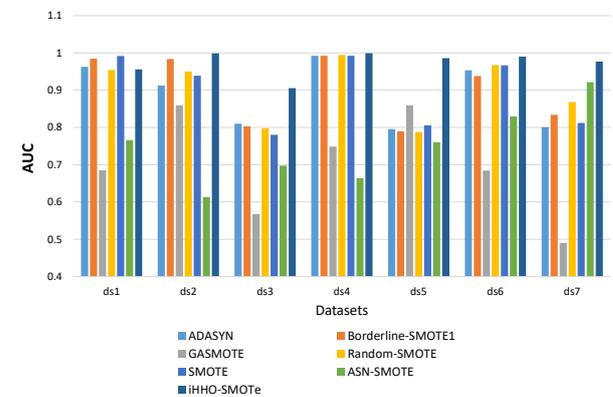

**Fig. 6 Comparison of AUC of seven SOMTE techniques**

This research uses a well-known dataset for detecting credit card fraud. The dataset was created by the ULB Machine Learning Group, which focuses on big data mining and fraud detection. It includes credit card transactions made by European clients over two days in September 2013. Out of 284,807 transactions, 492 were fraudulent.

Most of the attributes in the dataset are numerical, thanks to a dimensionality reduction technique called Principal Component Analysis (PCA). The only exceptions are the "Time" and "Amount" attributes. "Time" represents the seconds that have passed since the first transaction in the dataset, and "Amount" is the cost of the transaction. The "Class" attribute is the target variable, with a value of 1 indicating a fraudulent transaction and 0 indicating a legitimate one.

In Fig.7, compares different methods using this credit card fraud dataset, which is known for its large volume of transactions. The goal was to thoroughly evaluate the stability and accuracy of the new method, especially in the context of big data challenges, against other techniques. The results showed that iHHO-SMOTe achieved the highest AUC score of 0.968, outperforming other methods, which scored below 0.94. These other methods, ranked from highest to lowest AUC, were Borderline-2, SMOTE, ADASYN, Borderline-1, ASN-SMOTE, GASMOTE, and Random SMOTE.

In terms of the F1-score, all algorithms performed exceptionally well, with scores consistently above 0.99, and some even reaching a perfect score of 1. For the G-mean metric, iHHO-SMOTe again demonstrated its superiority with a score exceeding 0.97, while the other methods scored below 0.95 except borderline-2 which socres 0.95. This demonstrates iHHO-SMOTe's effectiveness and robustness in handling credit card fraud detection within large datasets

In Fig.8, compare between each of Gmean, F1-score, and AUC for SOMTe techniques before and after clearing noised and border. is it shows that Random smote have less G-mean and f1 and more enhancment in AUC and for Smote, ADASYN, borderlin1, borderlin2, GASMOTE, ANS-SMOTE, i-HHO-SMOTE have More G-mean and less f1 and AUC.

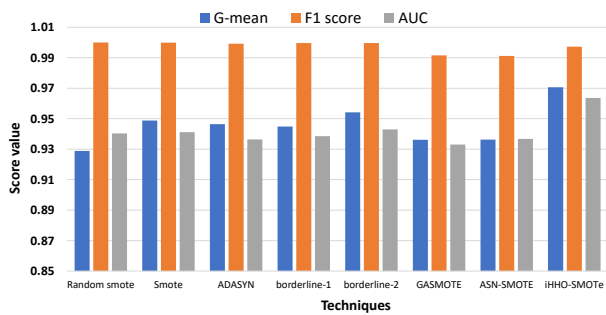

**Fig. 7 Comparison of SMOTE techniques and iHHO-SMOTe using fraud detection dataset.**

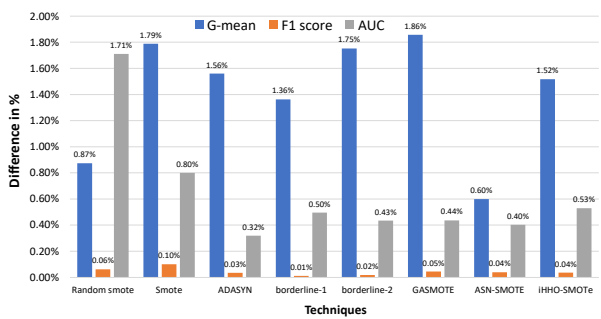

**Fig. 8 The difference between each of Gmean, F1-score, and AUC for SOMTe techniques before and after clearing noised and border.**

## 6 Conclusion

In summary, the iHHO-SMOTe approach is a big step forward in handling noised and border imbalanced datasets in classification tasks. By combining Random Forest and DBSCAN algorithms with the Harris Hawk optimization algorithm and SMOTE, we've created a strong framework that can make accurate and reliable predictions even when data noised and imbalanced. This has important implications for many real-world applications where better classification accuracy and balanced data are crucial for making good decisions. Additionally, this research makes a significant contribution to the field of imbalanced data handling. It introduces a powerful method that improves the performance of classification models across different areas. This combination of advanced techniques can effectively address the challenges of skewed data distributions, leading to more accurate and trustworthy predictions.